\documentclass[runningheads]{llncs}
\usepackage[T1]{fontenc}
\usepackage{wrapfig}
\usepackage{pgfplots}
\pgfplotsset{compat=1.16}

\usepackage{graphicx}
\usepackage{xcolor}

\usepackage{amsmath}
\usepackage{algorithm}
\usepackage{multirow}
\usepackage{algpseudocode}

\begin{document}
\title{Prophecy: Inferring Formal Properties from Neuron Activations}
%
%
\author{Divya Gopinath
\and
Corina P\u{a}s\u{a}reanu
\and
Muhammad Usman
}

%
\institute{NASA Ames, KBR, CMU} 
%
\maketitle              
\begin{abstract}
We present Prophecy, a tool for automatically inferring
formal properties of feed-forward neural networks. Prophecy is based on the observation that a significant part 
of the logic of feed-forward networks is captured in the activation status of the neurons at inner layers. Prophecy works by extracting rules based on neuron activations 
(values or \textit{on}/\textit{off} statuses)  as preconditions that imply certain desirable output property, e.g., the prediction being a certain class. These rules represent network properties captured in
the hidden layers that imply the desired output behavior. We
present the architecture of the tool, highlight its features and demonstrate its usage on different types of models and output properties. We present an overview of its applications,  such as inferring and proving formal explanations of neural networks, compositional verification, 
run-time monitoring, repair, and others. We also show novel results highlighting its potential in the era of large vision-language models.

\end{abstract}

\section{Introduction}

Deep Neural Networks (DNNs) are being used in many tasks, some of them in critical domains, such as banking, medicine, or transportation, raising safety and security concerns. These are due to many factors, among them lack of robustness and transparency and also lack of intent. In fact, neural networks only learn from examples, often without a high-level requirement specification. In contrast, more traditional safety-critical software systems are usually designed based on high-level, often formalized, requirements.

\begin{figure}
    \centering
    \includegraphics[scale=0.26]
    {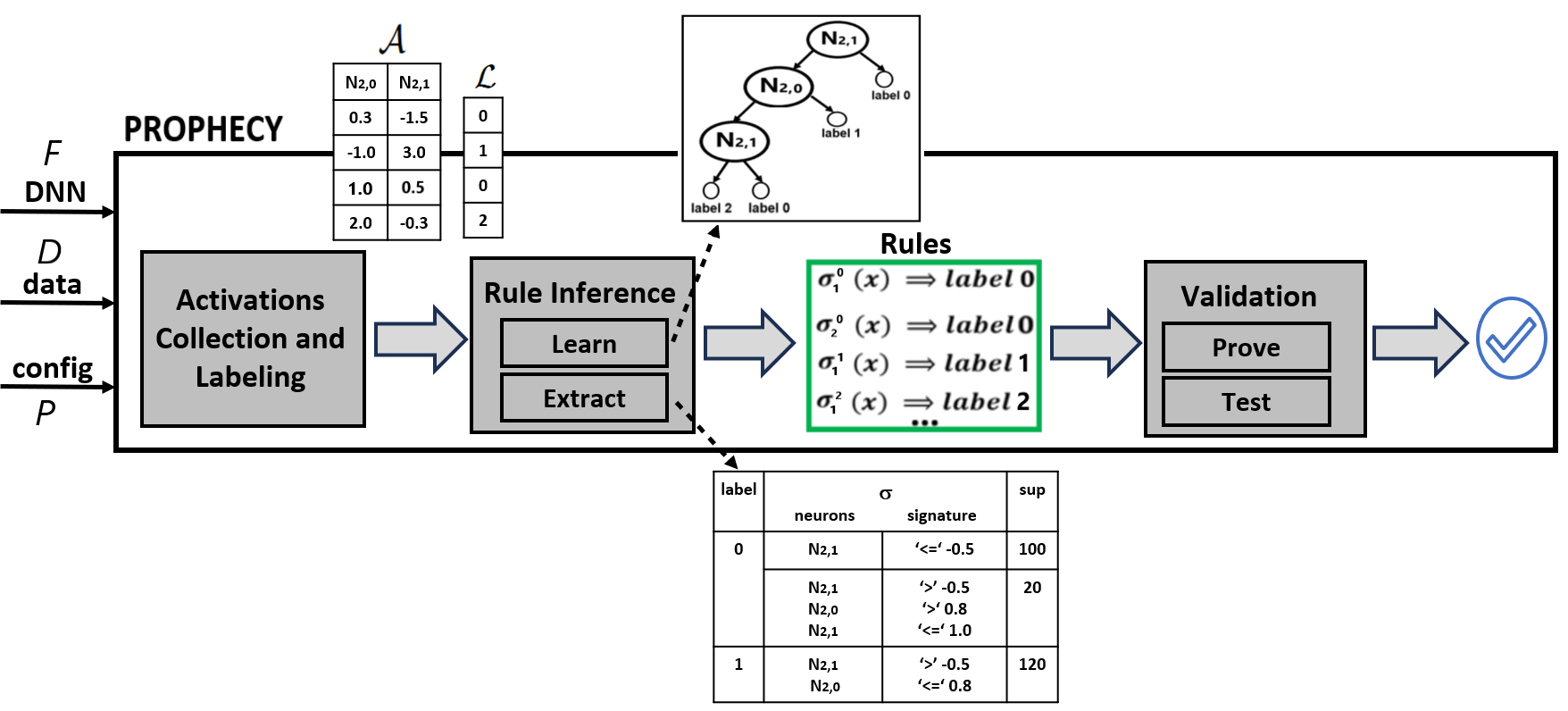}
    \caption{Prophecy Tool. Overview Diagram.}
    \label{fig:prophecy_overview}
\end{figure}

In this paper, we present \textbf{Prophecy}, a tool for automatically
inferring formal properties of feed-forward neural networks. Neural networks work by applying layer transformations to the inputs, extracting important features from the data, and making decisions based on these features. Prophecy aims to extract \textit{layer properties} that capture common characteristics over the extracted features, allowing insight into the inner workings of the network. These properties are of the form $Pre \Rightarrow Post$. $Post$ is a post-condition stating some desired output behavior. They are typically simple and easy to specify, for instance, the network's prediction being a certain class. $Pre$ is a precondition that Prophecy automatically infers and can serve as a formal explanation for why the output property holds. Each layer property groups inputs that have common characteristics observed at an intermediate layer and that together imply the desired output behavior. 

Several studies \cite{NIPS2012c399862d,understandingconvolutionalnetworks,FeatureVis}
have shown that neuron activations (values or \textit{on}/\textit{off} statuses)  at a layer effectively represent the features at that layer. Typically, dense layers closer to the output encompass the
logic relevant to model decisions.  
Guided by these insights, simple rules in terms of conditions on neuron outputs at these layers, have the potential to capture the logic corresponding to a desired model behavior. 
Prophecy uses a relatively inexpensive technique that relies on data to infer such \textbf{rules}. 
The idea is to observe neuron activations for a large number of inputs and learn patterns that act as the $Pre$ implying desired output properties. In particular, it uses \textit{decision-tree learning}~\cite{quinlan1986induction,quinlan1993c4}, an efficient algorithm which employs various information-theoretic measures to yield rules with simple logical constraints that are compact and thus generalizable.\\

\noindent \textbf{Methodology.}
The Prophecy framework (Fig.~\ref{fig:prophecy_overview}) takes as inputs; a pre-trained model ($F$), a post-condition or output property ($P$), and a set of data ($D$) representative of in-distribution for the model. 
The \textit{Activations Collection and Labeling} module evaluates the network on each input, $x \in D$, to collect the neuron values  or just the \textit{on/off} (1/0) activations at a given layer $l$. This creates the set of all neuron activations, $\mathcal{A}$. The property $P (F (x))$ is evaluated on each input to
yield a boolean value indicating whether the output $F (x)$ satisfies the property $P$ or not. A labels dataset ($\mathcal{L}$) is generated based on the evaluation of $P$ on $\mathcal{D}$. However, the actual label values in $\mathcal{L}$ are determined based on the type of post-condition property. For instance, when the output property is a prediction post-condition; $P (F(x)) ::= (F(x) = c)$, $\mathcal{L}$ is a list of the corresponding class values. This enables using the decision-tree to directly predict the class label rather than a boolean as to whether the prediction is $c$. 

The \textit{Rule Inference} module invokes the decision-tree learning algorithm to train an estimator using the datasets $\mathcal{A}$ and $\mathcal{L}$. The resulting tree predicts the  valuation of $P$ on any given input. Each path of the tree corresponds to a neuron pattern, $\sigma$; a conjunction of conditions in terms of the neuron activations. A path leading to a $True$ valuation of $P$ can be interpreted as a rule which states that for all inputs satisfying $\sigma$ at layer $l$, $P(F(x))$ would hold true. Thus, this module generates a collection of  \textit{layer properties}; each predicting model behavior w.r.t a desirable property in terms of a conditional logic on the features at that layer.
The \textit{Validation} module enables evaluating the generated rules and selecting those with high quality for subsequent applications. Each rule can be empirically validated on a held-out dataset to obtain confidence in its precision. The module also enables formal verification of the rules (using a decision-procedure $DP(F,\sigma,P,l)$) to provide guarantees about model behavior on all inputs satisfying the pre-condition.

Prophecy, thereby, serves to decompose a complex black-box DNN model, into a set of compact rules amenable to analysis. Rules act as likely specifications of the model with a precise mathematical form. This in turn enables multitude of applications; obtaining formal explanations and guarantees of behavior, testing, debugging, analysis in terms of human-understandable concepts and runtime monitoring. The idea, initially presented in \cite{ProphecyASE19}, is one of the first formal approaches to look into the black-box of a DNN model. Prophecy and its applications are well-referenced (over 55 citations) and explored both in the industry (NASA, Boeing, DoT, FAA) and academia (York, Stanford, McMaster University and University of Bergamo).\\

\noindent{\textbf{Related Work.}} Approaches such as~\cite{NNDTs,NNDistill,DecTreeExt,NDT} construct alternate, lightweight and explainable representations of a DNN; decision-trees that are functionally equivalent to the entire model. Prophecy, in contrast, focuses on using decision-tree learning to build a set of high-quality rules in terms of the features at an inner layer. In this sense, it is closer to \textit{probing}~\cite{ProbingANN,ProbClass}, a popular 
analysis strategy, most widely used for NLP tasks~\cite{Emergent,ProbSem,AlainB16}. However, these techniques mainly target interpretability; mapping activations to high-level concepts~\cite{KimWGCWVS18}, with a separate step (such as intervention~\cite{AmnesicProbe}) to establish causal relationship with model outputs. Prophecy offers the flexibility to vary the target of the probe to enable a variety of analyses. In contrast to all aforementioned methods, the 
scope is restricted to a specific output property and to the set of inputs satisfying the pre-condition. This enables building simple yet precise rules for analysis.\\

\noindent\textbf{Contributions.}
Prophecy has never been published as a tool and this work focuses specifically on this aspect to promote more wide-spread usage.

\begin{enumerate}
    \item \textit{Providing a functional tool:} The basic idea and methodology of Prophecy was presented in \cite{ProphecyASE19}; the prototype implementation required the user to manually edit the code to enable different functionalities (such as extracting different types of rules for different post-conditions for different models). In addition, the user had to manually process the extracted rules and invoke the solver externally to formally check them. In this work, we present a functional automated tool, \textit{DNN-Prophecy}\footnote{Subsequent references to Prophecy imply the new version unless explicitly specified.}. The implementation was refactored to modularize the code and enable invocation of different functionalities (including the invocation of Marabou~\cite{Marabou} solver) via command-line parameters. We have also extended the implementation to extract and prove more general rules (not just in terms of on/off activations).
    \item \textit{Optimizing implementation to improve efficiency:} We extend the previous implementation (\cite{ProphecyASE19}) to speed up accessing the rules at runtime (Section~\ref{sec:tool_eval}).
    \item We summarize the many applications of Prophecy since 2019, namely, repair, test coverage, runtime monitoring, poison-attack mitigation, and feature-guided analysis (Section~\ref{sec:applications}), highlighting the effectiveness of the layer properties which act as logical probes in the embedding space of the network.
    \item  We present a novel application and evaluation of Prophecy on a Vision Language Model (Section~\ref{sec:tool_eval}).
\end{enumerate}
\section{Tool Description}
In this section we describe in more detail the rules extracted by Prophecy, the algorithms employed and the tool usage.

\subsection{Definition of rules}

\begin{wrapfigure}{r}{0.63\textwidth}
\vspace{-5mm}
\centering
    \includegraphics[scale=0.15]{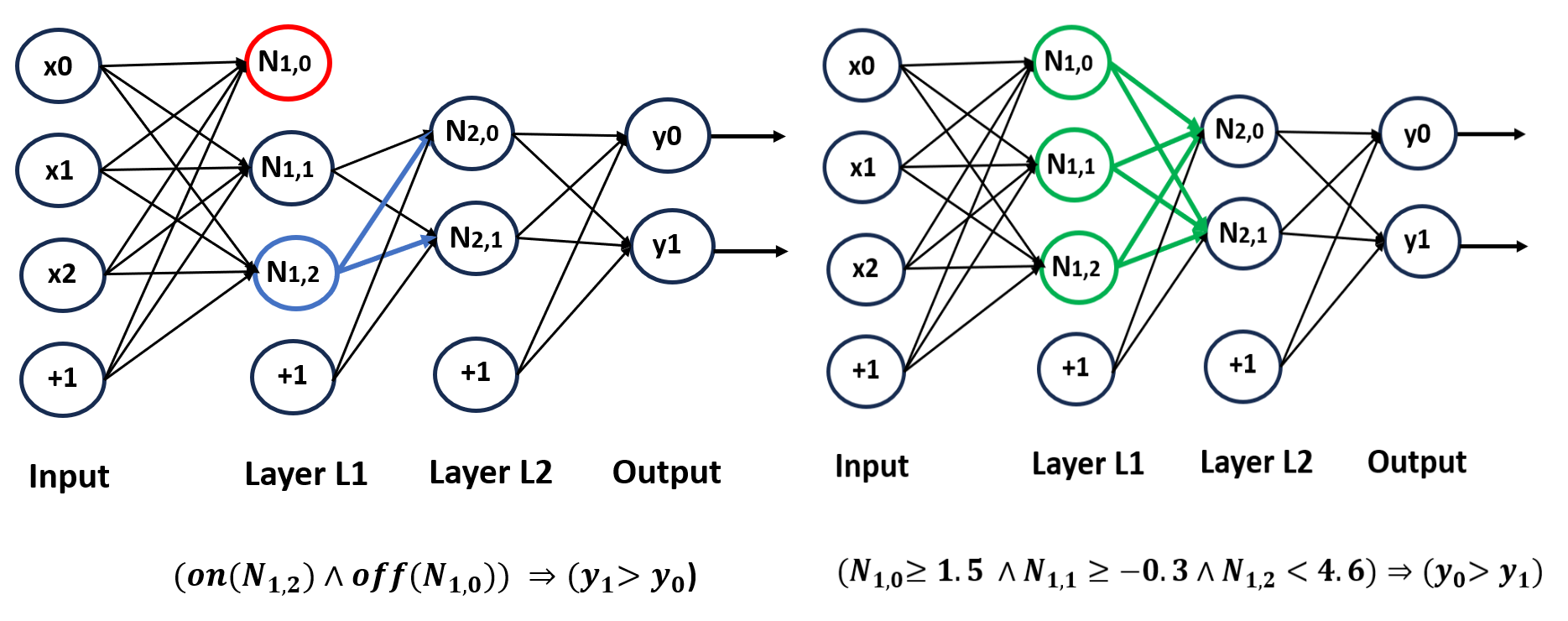}
    \vspace{-3mm}

    \caption{Example Layer Properties.}
    \label{fig:layer_properties}
    \vspace{-3mm}
\end{wrapfigure}
Let $F$ denote a feed-forward neural network and $F_{l}$ denote the function representing the transformations applied by all layers from the input till layer $l$. 
A layer property for a post-condition P encodes a pattern $\sigma$ over the neuron values in layer $l$ that satisfies the rule,  
$\forall x \in X \; \sigma(F_{l}(x)) \Rightarrow P (F (x))$.

In the original paper \cite{ProphecyASE19}, we expressed $\sigma$ in terms of \textit{on/off} activations, $\sigma(X) ::= \bigwedge_{N_{i} \in (N_{\sigma} \subseteq N_{l})} (\mathit{on}(N_{i}(X)) \lor \mathit{off}(N_{i}(X)))$, $\mathit{on}(N_{i}(X)) \iff (N_{i}(X) > 0)$,  $\mathit{off}(N_{i}(X)) \iff (N_{i}(X) <= 0)$. 
$N_{l}$ is the set of neurons at layer $l$, each of which can be recursively expressed as a function of $X$. $N_{\sigma}$ is a subset of $N_{l}$ constrained by $\sigma$. 
We have since then extended $\sigma$ to express more general mathematical constraints over the neuron values, $\sigma(X) ::= \bigwedge_{N_{i} \in (N_{\sigma} \subseteq N_{l})} N_{i}(X) \; op \; V_{i}$, where $V_{i}$ is a real-valued threshold, and operator $op \in \{ >=, <=\}$. The neuron pattern defines a convex region in $F_l(X)$ and a union of convex regions in the input space. 

Figure~\ref{fig:layer_properties} (left) shows an example property that specifies that for all inputs with $\mathit{on}(N_{1,2})$ and $\mathit{off}(N_{1,0})$ at layer $L1$ (and any value for other neurons), the output is most likely label 1 ($y_1 > y_0$). The other property (right) specifies conditions in terms of specific values which lead to label 0, ($y_0 > y_1$).

\subsection{Algorithms for Activations Collection, Labeling, and Proving}
\label{subsec:acts_col}
{\bf Activations Collection and Labeling} implements the functionality described in
Algorithm~\ref{alg:collect_label}, taking in the pre-trained model F, dataset D (a set of model inputs and corresponding labels) and configuration parameters ($C_{config}$) as set by the user. The parameter, $layer\_name$, specifies the layer at which the activations are to be collected and is typically one of the dense layers close to the output. The $acts$ parameter determines if on/off neuron activations are to be collected or neuron values. Note that the parameter $inptype$ can be set to 1 to optionally skip executing the model to collect the layer activations. In this setting, the user-provided dataset ($x$) is expected to contain the list of activation vectors (refer Section~\ref{subsec:reg}).

The purpose of the {\em LABEL} function is to evaluate the user-provided post-condition property, $P$, on the network output for each input. The $type$ parameter acts a short-cut for specifying standard properties on the predicted labels of typical classification models. The parameter when set to 0, enables evaluation of  $P(F(X)) ::= (F(X) = c)$. The label predicted by the model for each input is used to populate $\mathcal{L}$. The $type = 1$ indicates a correctness post-condition; $P(F(X)) ::= (F(X) = c_{ideal})$. The input labels dataset ($y$) is expected to contain the ideal ground-truth labels for every input and $\mathcal{L}$ is populated with either 1 or 0 based on the satisfaction or violation of correctness respectively. Similarly, $type = 2$, enables evaluation of $P(F(X)) ::= (F(X) = c) \wedge (F(X) = c_{ideal})$. 
Currently the tool enables handling models which are not classification and/or other post-condition properties by setting $type = 3$. In this setting, the input labels dataset ($y$) is expected to directly contain the results of the post-condition evaluation on the model output for each input (refer Section~\ref{subsec:reg} in the appendix).
\vspace{-4mm}
\begin{algorithm}
\caption{Activations Collection and Labeling}
\label{alg:collect_label}
\begin{multicols}{2}
\begin{scriptsize}
\begin{algorithmic}[1]
\Require \(F\),\( D ::= \{(x,y)\} \),
\( C_{\text{config}}::= \{type,\)
\;\;\;\;\;\;\;\;\;\;\( layer\_name,acts,inptype, ...\} \)
\Ensure \( \mathcal{A} \), \( \mathcal{L} \)
\State \( A \gets \emptyset \), 
\( L \gets \emptyset \), 
\( l \gets layer\_name \)

\For{\( x \in D \)}
    \If{$inptype = 0$}
        \If{$acts = \mathtt{False}$}
            \State $A \gets A \cup \{F_l(x)\}$ 
        \Else
            \If{$F_l(x) > 0$} $A \gets A \cup \{1\}$
            \Else \; $A \gets A \cup \{0\}$ \EndIf
        \EndIf
    \Else
        \State $A \gets A \cup \{x\}$ 
    \EndIf
    \State \( c \gets \text{LABEL}(x,y,F,type) \) 
    \State \( L \gets L \cup \{c\} \) 
\EndFor
\State \Return \( (A, L) \)
\State
\Function{label}{$x, y, F, \text{type}$}
\If{\(\text{type} = 0\)} \Return \(F(x)\)
\ElsIf{\(\text{type} = 1\)}
    \If{\(F(x) = y\)}  \Return 1
    \Else \; \Return 0
    \EndIf
\ElsIf{\(\text{type} = 2\)}
    \If{\(F(x) = y\)} \Return \(F(x)\)
    \Else \; \Return 1000
    \EndIf
\ElsIf{\(\text{type} = 3\)} \Return \(y\)
\EndIf
\State \Return \(\text{None}\)
\EndFunction
\end{algorithmic}
\end{scriptsize}
\end{multicols}
\vspace{-4mm}
\end{algorithm}

\vspace{-4mm}

\noindent\textbf{Rule Inference} first executes the {\em Learn} module, which invokes the \textit{sklearn. tree.DecisionTreeClassifier} library \cite{scikit} to build a decision-tree estimator. 
Each path of the tree from the root to a leaf forms a pattern (in terms of the values in $\mathcal{A}$) for predicting a label in $\mathcal{L}$. There can be more than 1 paths leading to the same label (example tree in  Fig.~\ref{fig:prophecy_overview}). Each path can also lead to a leaf with more than 1 labels (impure). 
The next step, $Extract$, parses the tree to collect paths leading to \textit{pure} leaves. This process builds a dictionary; mapping each distinct label in $\mathcal{L}$ to a set of prediction rules. Each rule is represented by a triple, \{neurons, signature, support\} (Fig.~\ref{fig:prophecy_overview}). 
Let \(\{\sigma^{c}_1, \sigma^{c}_2, \ldots, \sigma^{c}_n\}\) denote the set of rules associated with label \(c\), where each rule \(\sigma^{c}_i\) has an associated support metric, \text{supp}($\sigma^{c}_i$)\, defined as the number of instances in $D$ that satisfy $\sigma^{c}_i$ (and $P (F(X))$ since only pure rules are extracted). Each rule can be \textbf{validated} by the $\mathit{Test}$ module which computes statistical metrics (precision, recall and F1) on a separate dataset. 

\vspace{-4mm}
\begin{algorithm}
\caption{Iterative Rule-proving using Marabou~\cite{MarabouGITHUB}. The class and its methods from Marabou are highlighted in blue.}
\label{alg:mar_proof}
\begin{multicols}{2}
\begin{scriptsize}
\begin{algorithmic}[1]
\Require \(c\), \(l\),
$X^{s}_{\text{min}}$, $X^{s}_{\text{max}}$, $V^{s}_{\text{min}}$,$V^{s}_{\text{max}}$,
$X_{0}$,$V_{0}$,

 \;\;\;\;\(F_{onx}: \textcolor{blue}{MarabouNetworkONNX}\)
\Ensure \( result \), \( labs \)
\State \( result \gets False \), 
\( it \gets 0 \), \texttt{$labs$ $\gets$ \{$c' \neq c$\}}
\While{\texttt{$result = False$}}
    \If{\texttt{$it$ = 0}}
    \State \texttt{MB\_SETBNDS(}$X^{s}_{\text{min}}, X^{s}_{\text{max}}$,$V^{s}_{\text{min}}, V^{s}_{\text{max}}$,
    
    \;\;\;$l$,$F_{onx}$\texttt{)}
    \EndIf

    \If{\texttt{$it$ = 1}}
    \State \texttt{MB\_SETBNDS(}$X^{s}_{\text{min}}, X^{s}_{\text{max}}$,$V^{0}, V^{0}$,
    
    \;\;\;$l$,$F_{onx}$\texttt{)}
    \EndIf

    \If{\texttt{$it$ = 2}}
    \State \texttt{MB\_SETBNDS(}$X^{0}, X^{0}$,$V^{0}, V^{0}$,$l$,$F_{onx}$\texttt{)}
    \EndIf
    
    \State $result$, $labs$ = 
    
    \texttt{MB\_SOLVE($labs$,$F_{onx}$,$t$)}
    \State $it = it + 1$
\EndWhile
    \State \Return $result$, $labs$
\State
\Function{MB\_SETBNDS}{$X_{min}$,$X_{max}$,

$V_{min}$,$V_{max}$,$l$,$F_{onx}$}

\State $Ivars = F_{onx}.\textcolor{blue}{inputVars}$
\For{$idx = 0$ to $|Ivars|$}
    \State $i = Ivars[idx]$
    \State $F_{onx}.\textcolor{blue}{setLowerBound}(i,X_{min}[idx])$
    \State $F_{onx}.\textcolor{blue}{setUpperBound}(i,X_{max}[idx])$
\EndFor

\State $Lvars =$

\;\;\;\;\;$F_{onx}.\textcolor{blue}{layerNameToVariables[l]}$
\For{$idx = 0$ to $|Lvars|$}
    \State $i = Lvars[idx]$
    \State $F_{onx}.\textcolor{blue}{setLowerBound}(i,V_{min}[idx])$
    \State $F_{onx}.\textcolor{blue}{setUpperBound}(i,V_{max}[idx])$
\EndFor

\State \Return \( None \)
\EndFunction
\State
\Function{MB\_SOLVE}{$labs$,$F_{onx}$,$t$}
    \State $Ovars = F_{onx}.\textcolor{blue}{outputVars}$
    \State $labvar = Ovars[c]$
    \For{$idx = 0$ to $|Ovars|$}
    \State $v = Ovars[idx]$
    \If{ $idx = c$} 
    \State \textbf{continue} 
    \EndIf
    \State $F_{onx}.\textcolor{blue}{addConstraint}(labvar > v)$
    \State $res,vals,stats =$ 
    
    \State $F_{onx}.\textcolor{blue}{solve}(timeout=t)$
    
    \If{\texttt{$res$ = 'sat' \textbf{or} 'TIMEOUT'}}
    \State \Return \texttt{$False$, $labs$}
    \Else \;\;\texttt{$labs$ $\gets$ $labs - \{idx\}$}
    \EndIf
    \EndFor
    
\State \Return \(True, \emptyset\)
\EndFunction

\end{algorithmic}
\end{scriptsize}
\end{multicols}
\vspace{-4mm}
\end{algorithm}

\vspace{-4mm}

Each rule can be formally verified by the \textbf{Prove} module. Given a label, $c$, the rule with the highest support ($\sigma^{c}_{s}$) is selected;
$\sigma^{c}_{s} = \underset{\sigma^{c}_i}{\arg\max} \; \text{supp}(\sigma^{c}_i)$, $l$ refers to the corresponding layer. This rule has the highest chance of satisfying the verification check; $\forall x : \sigma(F_{l}(x)) \Rightarrow P (F (x))$. In order to characterize the region satisfying the pre-condition, 
over-approximation boxes encompassing the inputs in $\sigma^{c}_{s}$ are computed in the input space (dimensionality $I$) and in the latent space (dimensionality $L$) respectively.
{\footnotesize
$$X_{\text{sup}} = \{ x \in {D} \mid \sigma^{c}_{s}(x) \},
X^{s}_{\text{min}} = \big[\min \limits_{x \in X_{\text{sup}}} x_i \big]_{i=0}^{I-1},  X^{s}_{\text{max}} = \big[\max \limits_{x \in X_{\text{sup}}} x_i \big]_{i=0}^{I-1},
$$

$$V^{s}_{\text{min}} = \big[\min \limits_{x \in X_{\text{sup}}} F_{l}(x) \big]_{i=0}^{L-1}
, V^{s}_{\text{max}} = \big[\max \limits_{x \in X_{\text{sup}}} F_{l}(x) \big]_{i=0}^{L-1}
$$
}

Let us consider the prediction post-condition of typical classification models, $P(F(X)) ::= (F(X) = c)$. Algorithm~\ref{alg:mar_proof} shows the algorithm invoking Marabou \cite{Marabou} solver to perform the check \footnote{Although the current implementation uses Marabou, it is extensible to other solvers.}. The solver takes in models in the ONNX format and the algorithm invokes the methods in the \textit{MarabouNetworkONNX}~\cite{MarabouGITHUB} API. 

In the first iteration, the network input and layer variables are bound to $X^{s}_{\text{min}}$, $X^{s}_{\text{max}}$, $V^{s}_{\text{min}}$, $V^{s}_{\text{max}}$ respectively. The function, $MB\_{SOLVE}$, is called to check the output constraints $\forall c' \neq c \; (F_{logits}(x)[c] > F_{logits}(x)[c'])$. The output variables are constrained to represent the negation of $P$; (i.e.) for every $c' \neq c$ following query is formulated,
$\exists x \mid 
(F_{logits}(x)[c'] > F_{logits}(x)[c])$. The solver is invoked for the query corresponding to each label in $labs$ (which contains all $c' \neq c$ in the first iteration). A SAT solution indicates a counter-example; the presence of an input which satisfies $\sigma$ for which the network produces a different label ($c'$). An UNSAT indicates proof for the query w.r.t. the specific label, $c'$, and the label is removed from $labs$. The entire proof goes through ($result = True$) when the solver returns UNSAT for all labels in $labs$. In the case of SAT solutions for some queries, or a timeout, the result is set to $False$ and the updated $labs$ set is used in the next iteration. Subsequent iterations further restrict first the layer variables and then the input variables to values corresponding to a single satisfying instance ($V^{0}$, $X^{0}$). This ensures that the last iteration is guaranteed to be an UNSAT, which guarantees algorithm termination. The algorithm complexity is linear in the order of the number of classes. 

To summarize, the above algorithm first attempts to prove the rule on the entire region characterized by $\sigma^{c}_{s}$. If this step fails, it attempts to refine the region until the proof goes through. Note that that the proof may go through for a subset of queries, providing robustness w.r.t. mis-classification to specific labels (refer Section~\ref{sec:tool_usage} for an example). For other post-condition properties including those corresponding to non-classification models, the tool can take in a constraints file with user-provided constraints (Section~\ref{subsec:reg} in the appendix).

\begin{table}
\centering
\begin{scriptsize}
\begin{tabular}{|*{10}{c|}}
\hline
\multirow{1}{*}{\textbf{Layer}}  & \multirow{1}{*}{\textbf{Neurons}} & \multirow{1}{*}{\textbf{Signature}} & \multirow{1}{*}{\textbf{Support}} & \multirow{1}{*}{\textbf{Label}}\\

\hline


dense\_14&[14,23,2,17,20,21,9,1,&&&\\
&31,12,13,27,4,29,3]&[0,0,0,0,0,1,0,1,0,1,0,1,1,1,1]&917&2\\
\hline

dense\_15&[1,0,7,6,3,2,8,&[$<=$1237,$<=$888.6,$<=$1388.8,$<=$1197.4,&&\\
&9,4,9,3,5]&$<=$1778.3,$<=$981.6,$<=$1232.2,$<=$2709.8,&&\\
&&$<=$1742.2,$<=$2664,$>$-2222.5,$>$2202.6]&4332&5\\

\hline
{dense\_14}&[2,7,4,7,1,1,5]&[$<=$5.02,$<=$5.23,$>$0.7,$<=$3.86&&\\
&&$<=$-1.43,$<=$-1.73,$>$,-3.6]&54&1000\\

\hline
\hline

\textbf{Train} & \textbf{Train} & \textbf{Test} & \textbf{Test} & \textbf{Test}  \\
\textbf{Recall} & \textbf{F1} & \textbf{Precision} & \textbf{Recall} & \textbf{F1}  \\

\hline
15.66\%&27.09&94.66\%&18.44\%&31.11\\
\hline
80.86\%&89.42&100\%&78.18\%&87.75\\
\hline
50.29\%&66.92&80\%&30\%&43.64\\
\hline

\end{tabular}
\end{scriptsize}
\caption{Example rules and their metrics produced by the \textit{analyze} command. Train Precision is 100\% for all rules.} 
\label{tab:example_ops}
\end{table}
\vspace{-10mm}

\subsection{Tool Usage}
\label{sec:tool_usage}

The tool can be found in the repository https://github.com/safednn-nasa/DNN-Prophecy.
It is implemented in Python 3.10. Prophecy can be invoked using three main commands; (i) \textit{analyze} , (ii) \textit{prove}, and (iii) \textit{monitor}. The \textit{analyze} command enables the execution of the activations collection and labeling, learn, extract, and test modules, and the \textit{prove} command enables the execution of the prove module. The \textit{monitor} command enables the use of the extracted rules for the evaluation of the model on new inputs at inference time. Table \ref{tab:commands} (in the appendix) lists Prophecy commands and their respective parameters. We illustrate below an example usage of the tool. We provide more examples, including application of the tool on a regression model in the appendix.\\

\noindent{\textbf{Prophecy on MNIST classification model:}} Let us consider a model trained on the standard MNIST dataset~\cite{MNISTWebPage} to classify images into digits 0 through 9. We consider a simple model with two convolutional layers followed by max pooling, flattened into two fully connected dense layers ($dense\_14$ with 32 neurons, $dense\_15$ with 10 neurons). We would like to infer rules that imply classification to each unique output digit and provide formal guarantees\footnote{Refer /examples/Prophecy\_Tool\_MNIST.ipynb in the repository}. 

Prophecy is first invoked using the  \textit{analyze} command as shown below\footnote{PROPHECY\_PATH refers to the local path to the DNN-Prophecy repository.}. The desired post-condition is $(F(X) = c) \wedge (F(X) = c_{ideal})$; i.e. we would like rules that characterize inputs the model correctly classifies to every digit. This is enabled by setting the $type = 2$. The standard train datasets are used to obtain the input images and ground-truth labels to learn the rules.

\begin{small}

\noindent{\verb|$ python -m prophecy.main analyze|}

\verb|-m 'PROPHECY_PATH/dataset_models/cnn_max_mnist2.h5'|

\verb|-wd 'PROPHECY_PATH/results/mnist/rules/'|

\verb|-tx ./x_train.npy -ty ./y_train.npy -vx ./x_val.npy -vy ./y_val.npy|

\verb|-type 2 -acts True -layer_name 'dense_14'|

\end{small}

The rules are in terms of the (\textit{on/off)}) neuron activations at the dense layer, $dense\_14$. Note that using $odl$ , $oal$ will consider every dense and activation layer in the model respectively. The tree estimator and the rules extracted from it are stored in the working directory specified by $wd$. There are one or more rules for correct classification to each label and rules for incorrect classification where $c \neq c_{ideal}$ (label 1000) respectively. Each rule is statistically validated on the provided validation datasets. Table~\ref{tab:example_ops} shows example rules and their metrics on train and validation sets. The second and third rules are in terms of neuron values, generated by invoking \textit{analyze} without setting $acts$ to $True$.

The next invocation using the \textit{prove} command (shown below) enables formal verification of the extracted rules. 

\begin{small}

\noindent{\verb|$ python -m prophecy.main prove|}

\verb|-wd 'PROPHECY_PATH/results/mnist/rules/'|

\verb|-mp '/usr/local/lib/python3.11/dist-packages/maraboupy/'|

\verb|-onx 'PROPHECY_PATH/dataset_models/cnn_max_mninst2.onnx'|

\verb|-onx_map 'PROPHECY_PATH/dataset_models/MNIST_H5_ONNX_MAP.csv'|

\verb|-tx ./x_train.npy -label 5 -pred True|

\end{small}

This command invokes Marabou from the build folder provided in $mp$ and uses the $onnx$ version of the CNN model. It requires an $onx\_map$ which maps the layer names in the keras and $onnx$ versions of the model. This example command verifies a rule for $label\;5$. The rule with the highest train support ($\sigma^{5}_{s}$) is selected (across all layers). The images in  $x\_train$ dataset are used to determine $X_{sup}$, which in turn helps constrain the input and layer variables during verification. The parameter $pred$ when set to \textit{True} acts as a short-cut to enable checking classification constraints on the model output; (i.e.)  output node 5 has the maximum value for all inputs satisfying $\sigma^{5}_{s}$. 

The second example rule in Table~\ref{tab:example_ops} for label 5 gets proved in 15 seconds in the first iteration; 

$\forall x \in [X_{\min}, X_{\max}],\; F_l(x) \in [V_{\min}, V_{\max}]:$ 

$\forall c' \in \{0,1,2,3,4,6,7,8,9\}:$ 
$(F_{logits}(x)[c'] > F_{logits}(x)[c])$

For the first example rule for label 2, the proof goes through w.r.t. six labels (4 thru 9) in the first iteration, while the solver times out for queries corresponding to the remaining labels. The entire process takes 20 minutes. 
\section{Applications of Prophecy}
\label{sec:applications}
\noindent{\textbf{Robustness Proofs.}}
Provably-correct properties (at some layer $l$) defined with respect to some output class characterize regions in the input space in which the network is guaranteed to give the same class, i.e. the network is {\em robust} \cite{ProphecyASE19}. Counterexamples generated from failed proofs represent potential adversarial examples, as they are close (in the representation space at layer $l$) to (regions of) inputs that are classified differently.


\noindent{\textbf{Compositional Verification.}}
 Deep networks deployed in safety-critical contexts, such as ACAS-Xu \cite{ProphecyASE19}, have properties of the form $A \Rightarrow B$, stating that for all inputs $x$ satisfying $A(x)$, the corresponding output $y$ ($= F(x)$) satisfies $B(y)$. 
 Proving such properties for multilayer feed-forward networks is computationally expensive. 
 The Prophecy-inferred patterns can be used to decompose proofs of such properties. 
 Given a layer pattern $\sigma$ at some middle layer $l$, the proof of $A \Rightarrow B$ can be decomposed into $A \Rightarrow \sigma$ and $\sigma \Rightarrow B$. 
 We first identify a layer pattern $\sigma$ that implies the output property $B$, and then prove $A \Rightarrow \sigma$ on the smaller network up to layer $l$. Furthermore, once a layer pattern $\sigma$ is identified for a property $B$, it can be reused to prove other properties involving $B$. 
 

\noindent{\textbf{Formal Explanations.}}
 Prophecy can also be used to infer {\em formal explanations} of DNN behavior \cite{ProphecyASE19}.  
 Every concrete input defines an assignment to the input variables $x_1 = v_1 \wedge x_2 = v_2 \wedge .. \wedge x_n = v_n$ that satisfies a layer rule $\sigma$. The problem is
to find  a minimal subset of the assignments such
that $x_{k_1} = v_{k_1} \wedge x_{k_2} = v_{k_2} \wedge.. \wedge x_{k_m} = v_{k_m} \Rightarrow \sigma$. 
We adopt a greedy approach that eliminates constraints iteratively and stops when $\sigma$ is no longer implied; the checks are performed with a decision procedure such as Marabou. The resulting constraints are network preconditions that formally guarantee the postcondition corresponding to $\sigma$.

\noindent{\textbf{DNN Repair.}}  NNrepair \cite{dnnrepair} is a constraint-based technique for repairing DNNs. The framework can fix the network at an intermediate layer or at the last layer. NNrepair uses Prophecy to infer correctness specifications to act as an oracle for repair. NNrepair identifies a set of suspicious neurons and incoming suspicious edges. It adds $\delta$ values that are $0$ in the concrete mode but symbolic in symbolic mode. NNrepair executes the DNN concolically along positive and negative examples. It then collects the values of suspicious neurons/edges in terms of symbolic expressions. NNrepair then adds a set of repair constraints which are then solved using an off-the-shelf solver.


\noindent{\textbf{Feature-guided Analysis of Neural Networks.}}
In~\cite{Feature-Guided-DNN} we describe how to employ Prophecy to extract high-level, human-understandable features from the neural network internal representation, based on monitoring the neural network activations, and using high-level features instead of classes as postconditions. 
The extracted rules can serve as a link to high-level requirements and can be leveraged to enable various software engineering activities, such as automated testing, debugging, requirements analysis, and formal verification, leading to better engineering of neural networks. 



\noindent{\textbf{{Run-time Monitoring.}}
Prophecy can also be used to determine reliability of DNNs in deployment \cite{RuntimeMisclass}. In this work we explore the use of Prophecy rules in predicting, at runtime, whether the output of a network is correct or not. 
We use a set of passing and failing data to mine rules for correct and incorrect classification. 
 At inference time, Prophecy evaluates a new, unseen input against the set of extracted rules. 
 Thus, Prophecy can be used to predict if the output of the DNN is correct or not in deployment, benefiting safety-critical applications.

\noindent{\textbf{{Mitigation against Poison Attacks.}}
AntidoteRT~\cite{antidotert} is another run-time approach that leverages Prophecy, this time to protect DNNs against poison or backdoor attacks. AntidoteRT uses Prophecy to extract neuron patterns by running a DNN on clean and poisoned samples. These patterns are used at run time, to detect whether unseen inputs are likely poisoned or not. The patterns are also used as oracles to guess the ideal label for the detected poisoned inputs.

\noindent{\textbf{Rule-Based Test Coverage.}}
Prophecy is also useful in improving the testing of DNNs \cite{rule-based-testing}.
The goal is to use the rules to automatically generate ground truth for unlabeled data and to provide a new coverage metric which can evaluate existing test suites in terms of their functional diversity, defect detection ability, and sensitivity to different input scenarios. 


\section{Tool Evaluation}
\label{sec:tool_eval} 

This section describes and evaluates (i) a new feature that optimizes runtime monitoring, and (ii) a novel application of the tool on a large vision-language model.\\

\noindent{\textbf{Evaluation of Prophecy for Runtime Monitoring.}}
Prophecy provides functionality for runtime monitoring of the extracted properties~\cite{RuntimeMisclass} as outlined in Section~\ref{sec:monitor}. 
The \textit{monitor} command is used at inference time to tag the model's behavior on new inputs as \textit{correct}, \textit{incorrect} or \textit{uncertain}. In the previous implementation of Prophecy~\cite{ProphecyASE19}, after the inference of rules via decision-tree learning, the extract module was executed by default, which parsed the tree to create and store a set of rules. We have extended the implementation to have the decision-tree estimator stored as well in pickle format. Invoking the \textit{monitor} command using the \textit{classifiers} parameter, loads the estimator and uses it to directly \textit{predict} the model's behavior on the runtime inputs. This is much more efficient than a search through all the rules and therefore speeds-up the process.

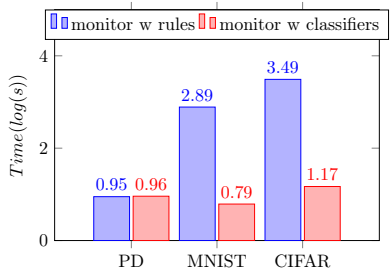
\begin{wrapfigure}{r}{0.47\textwidth}
\vspace{-5mm}
  \pgfplotstableread[row sep=\\,col sep=&]{
    interval & prophecy & prophecyplus  \\
    PD     & 0.95  & 0.96    \\
    MNIST     & 2.89 & 0.79    \\
    CIFAR    & 3.49 & 1.17  \\
    }\mydata

\begin{tikzpicture}[scale=0.68]
    \begin{axis}[
            ybar,
            bar width=0.7cm,
           enlarge x limits={abs=1.5cm}, 
           enlarge y limits=false,
            width=.65\textwidth,
            height=.5\textwidth,
            legend style={at={(0.5,1)},
                anchor=north,legend columns=-1},
            symbolic x coords={PD,MNIST,CIFAR},
            xtick=data,
            nodes near coords,
            nodes near coords align={vertical},
            ymin=0,ymax=5,
            ylabel={$Time (log(s))$},
        ]
        \addplot table[x=interval,y=prophecy]{\mydata};
        \addplot table[x=interval,y=prophecyplus]{\mydata};
        \legend{monitor w rules, monitor w classifiers}
    \end{axis}
\end{tikzpicture}
    \caption{Inference times comparison. Pima Diabetes (PD~\cite{PD})), MNIST~\cite{MNISTWebPage}, CIFAR~\cite{CIFAR-10}.}
    \label{fig:graph}
\vspace{-10mm}
\end{wrapfigure}

The command when invoked with the  \textit{rules} parameter performs a linear search on the dictionary of rules (simulating original Prophecy implementation). Figure~\ref{fig:graph} compares the application of the \textit{monitor} command with \textit{rules}  vs. with  \textit{classifiers} . It highlights that using the classifier directly provides considerable savings in inference times (69.59\% on average) specifically for image models. Refer Section~\ref{sec:monitor} in the appendix for more details.\\
\noindent{\textbf{Evaluation of Prophecy on Large Models.}}
Recent work \cite{massive} indicates that there are certain activations (\textit{massive activations}) in large language models (LLMs) and vision language models (VLMs) that exhibit significantly higher values compared to other activations in the model. These activations are closely related to self-attention, causing the attention mechanism in these models to focus on the tokens associated with them. 
This motivated us to investigate the application of Prophecy to VLMs and see if Prophecy can scale and infer rules from a VLM. VLMs such as CLIP~\cite{pmlr-v139-radford21a} have a more complex architecture than traditional CNNs; with separate encoders for images and text respectively. They are trained on image-text pairs and can be used to select the caption (from a set of captions) that has the highest similarity to a given image. This strategy can be leveraged for {\em zero-shot classification} of images~\cite{pmlr-v139-radford21a}. We considered the OpenAI-ViT-Large-Patch14\footnote{https://huggingface.co/openai/clip-vit-large-patch14} CLIP model and a dataset of 5173 images (and respective ground-truth labels) of \textit{trucks} and \textit{cars} from the RIVAL-10 dataset. The zero-shot classification accuracy of CLIP on this data set was 98. 12\% (car: 98.46, truck:
97.77). 

Our goal was to extract rules for correct and incorrect zero-shot classification. We considered \textit{LayerNorm2} of the  ViT-L/14 vision transformer, which appears after the Multi-Head Self-Attention and before the MLP in every block, and has a dimensionality of (257 X 1024)\footnote{for a 224×224 image with 14×14 patch size}. We considered the $18^{th}$ layer of the encoder (as in the previous study on massive activations \cite{massive}). We invoked the {\em analyze} command of Prophecy with $inptype = 1$, which allows for activations vectors to be directly provided as input (skipping the need to execute the model to collect the activations). This facilitates the application of the tool to analyze different model architectures.
We considered 400 correctly classified and 97 mis-classified images from the original dataset. We obtained the neuron activations (external to Prophecy) using a \textit{forward hook}
; a 263,168 sized activation vector for each input. The activations were extracted within 262 seconds. A labels dataset was created (external to Prophecy) with binary labels corresponding to correct and incorrect zero-shot classifications respectively. Prophecy was then invoked to directly extract rules from the activations and labels datasets.
\begin{small}

\noindent{\verb|$ python -m prophecy.main analyze -type 3 -inptype 1|}

\verb|-tx features_train.npy -ty labs_train.npy|

\verb|-vx features_test.npy -vy labs_test.npy|

\end{small}
The rules in Table~\ref{tab:vlm1} (extracted in 81 seconds) show potential in capturing the logic that impacts VLM behavior. We explored another method to potentially improve the quality of the rules (Appendix~\ref{sec:vlm2}). This paves way to enabling different types of applications and analyses (Section~\ref{sec:applications}) to VLMs. 

\begin{table}[t]
\begin{scriptsize}
\centering
\begin{tabular}{|*{11}{c|}}
\hline
\multirow{2}{*}{\textbf{Label}} & \multirow{2}{*}{\textbf{Neuron}}	& \multirow{2}{*}{\textbf{Sign.}}	& \multirow{2}{*}{\textbf{Supp.}} & 	\textbf{Train}	& \textbf{Train}	& \textbf{Train}	& \textbf{Test}	& \textbf{Test}& 	\textbf{Test}& 	\multirow{1}{*}{\textbf{Avg}} \\ 

&&&&\textbf{Prec}&\textbf{Recall}&\textbf{F1}&\textbf{Prec}&\textbf{Recall}&\textbf{F1}&\textbf{F1} \\ \hline
\multirow{3}{*}{0}	& 139445  	&  $\leq$-1.788  & 	\multirow{3}{*}{21} & \multirow{3}{*}{100.00}	& \multirow{3}{*}{43.75}	& \multirow{3}{*}{60.87}	& \multirow{3}{*}{36.36}& 	\multirow{3}{*}{8.16}& 	\multirow{3}{*}{13.33}& 	\multirow{3}{*}{37.10}  \\  
& 100732&>0.415 &&&&&&&& \\
& 72927 &>-1.138 &&&&&&&& \\ \hline 

\multirow{6}{*}{1}	& 139445 	&  >-1.788 & 	\multirow{6}{*}{277} & \multirow{6}{*}{100.00}	& \multirow{6}{*}{92.33}	& \multirow{6}{*}{96.01} & \multirow{6}{*}{66.99}& 	\multirow{6}{*}{69.00}& 	\multirow{6}{*}{67.98}& 	\multirow{6}{*}{82.00} \\ 
& 52507 &$\leq$3.876 &&&&&&&& \\
& 75351 &$\leq$4.957&&&&&&&& \\
& 88245&>-4.015&&&&&&&& \\
& 135012&$\leq$6.902&&&&&&&& \\
& 174394 &>-3.071&&&&&&&& \\ \hline 

\end{tabular}
\caption{Rules for correct (1) and incorrect (0) zero-shot classification by CLIP. }
\label{tab:vlm1}
\end{scriptsize}
\vspace{-9mm}
\end{table}

\section{Conclusion} We presented Prophecy, a tool for inferring properties of DNN models. We described its applications and also reported on our initial investigation of Prophecy in the context of large (vision-language) models. In future work, we plan to explore in depth the many applications of Prophecy (as outlined at the beginning of Section~\ref{sec:applications}) but in the context of large language and vision-language models. 

\section*{Acknowledgments} We would like to thank our many collaborators on the Prophecy tool, in particular, Eduard Pinconschi, Andrew Wu, Ankur Taly, Hayes Converse, Huafeng Yu, Guy Katz, Burak Kadron, and Ravi Mangal.

\newpage

%
%
%
\bibliographystyle{splncs04}

\bibliography{main}

\newpage
\section*{Appendix}

\subsection{Tool Commands and Parameters.}
\label{sec:tool_com}
\begin{table*}[h!]
\begin{scriptsize}

\begin{tabular}{|c|l|l|}
\hline

\textbf{Command} & \textbf{Parameters} & \textbf{Description} \\\hline
\multirow{2}{*}{ALL} &\textit{-m}& Pre-trained model in keras (.h5) format. \\  \cline{2-3}
& \multirow{1}{*}{\textit{-wd}} & Working directory path\\ 
\hline
\hline
\multirow{25}{*}{\textit{analyze}}& \textit{-tx},\textit{-ty}& Data and labels (.npy) (\{x,y\} in Algorithm~\ref{alg:collect_label}).\\ \cline{2-3}
& \textit{-vx}, \textit{-vy}& Data and labels (.npy) for statistical validation.\\ \cline{2-3}

& &Layer/s to be used for activations collection.  \\ 
& \textit{-odl}&  \textit{-odl}: only dense layers (name starting with text 'dense'). \\
& \textit{-oal}&  \textit{-oal}: include the activation layers associated with dense layers. \\
& \textit{-layer\_name} & \textit{-layer\_name} (a specific layer).\\ \cline{2-3}

& \multirow{3}{*}{\textit{-inptype}}&Type of input data. (provided in \textit{-tx} and \textit{-vx}).   \\ 
& & \textit{0}: inputs to the model (eg. images).  \\
& & \textit{1}: array of neuron activations.\\ \cline{2-3}

& \multirow{5}{*}{\textit{-type}} & Short-cut for post-condition properties.\\ 
& & \textit{0}: rules w.r.t model output, eg. rules for every predicted label. \\
& & \textit{1}: rules for correct vs incorrect classification.\\
& & \textit{2}: rules for correct classification per label\\
& &  and incorrect classification. \\
& & \textit{3}: rules w.r.t labels in \textit{-ty} \\ \cline{2-3}

 & \multirow{3}{*}{\textit{-acts}} & Mathematical form of the rules.\\
 & & \textit{True}: in terms of \textit{on/off} neuron activations. \\ 

 & & In terms of neuron values if this parameter is not set to \textit{True}. \\ \cline{2-3}

& \multirow{3}{*}{\textit{-top}} & Number of rules.\\
& & \textit{True}: only rules with the highest train recall to be obtained. \\ 

&& All rules are obtained otherwise. \\ \cline{2-3}

 & \multirow{3}{*}{\textit{-sr}} & Option to skip rule extraction and statistical validation.   \\ 
&  & The decision-tree estimators are stored as is. \\
& & Saves memory space.\\ \cline{2-3}

& \textit{-b} & Balance classes in the train data. \\ \cline{2-3}
& \textit{-c}  & Consider model output confidence during rule extraction.\\ \cline{2-3}
& \textit{-rs} & Set the random state for reproducibility.\\

\hline
\hline
\multirow{13}{*}{\textit{prove}} & \multirow{1}{*}{\textit{-mp}} & Path to the Marabou build folder. \\ \cline{2-3}
&\multirow{3}{*}{\textit{-onx}} & ONNX model name corresponding to the keras model.\\ 
& & Marabou API takes models in .nnet or .onnx format.\\
& & Prophecy uses the ONNX API, the more widely used one.\\\cline{2-3}
& \multirow{3}{*}{\textit{-onx\_map}} & Maps layer names in keras and onnx models.\\
& & If not specified, Prophecy assumes the names to be the same. \\
& & The layer corresponding to the highest support is chosen.\\\cline{2-3}
& \textit{-label} & The label whose rules are to be considered for proving.\\ \cline{2-3}
& \multirow{2}{*}{\textit{-pred}} & True: Classification output constraints. \\ 
& & False: Non-classification output constraints. \\ \cline{2-3}
& \multirow{2}{*}{\textit{-min\_const}} &  Classification output constraints specifying that $label$ has\\
& &  the minimum value.\\ \cline{2-3}
& \multirow{2}{*}{\textit{-cp consts\_file}} & Path to a file specifying the  \\ 
& &  output constraints (ex. regression outputs).\\\cline{2-3}
& \textit{-tx}& data used to constrain network input variables.\\
& \textit{-tx}, \textit{-vx} & in-distribution data to calculate coverage of regions with proofs.\\
\cline{2-3}

\hline 
\hline
\multirow{3}{*}{\textit{monitor}}& \textit{-tx}, \textit{-ty} & Represents the unseen data and their ground-truth labels. \\ \cline{2-3}
& \textit{-rules} & Use the dictionary of rules to predict classes on unseen data. \\  \cline{2-3}
& \textit{-classifiers} & Use the tree estimator to predict classes on unseen data. \\ \hline

\end{tabular}

\label{tab:commands}
\end{scriptsize}
\caption{Prophecy Commands and Parameters.}
\label{tab:commands}
\vspace{-10mm}
\end{table*}

\subsection{Prophecy on the ACASXU classification model:}
\label{sec:ACASX}
ACASXU is a family of networks for collision avoidance system for unmanned aircraft control~\cite{ACASXU}. The model receives sensor information regarding the drone (the \emph{ownship}) and any nearby intruder drones, and then issues horizontal turning advisories aimed at preventing collisions. The input sensor data includes:
(1) Range: distance between ownship and intruder;
(2) $\theta$: angle of intruder relative to ownship heading direction;
(3) $\psi$: heading angle of intruder relative to ownship heading direction;
(4) $v_{\text{own}}$: speed of ownship;
(5) $v_{\text{int}}$: speed of intruder;
(6) $\tau$: time until loss of vertical separation; and
(7) $a_{\text{prev}}$: previous advisory.
The five possible output actions are as follows: (0) Clear-of-Conflict (COC), (1) Weak Left, (2) Weak Right, (3) Strong Left, and (4) Strong Right. Each advisory is assigned a score, with the \textit{lowest score} corresponding to the best action. 
The network used consists of 6 hidden layers, and 50 ReLU activation nodes per layer. 
The train dataset consists of 288165 inputs and the test dataset consist of 96056 inputs with corresponding known output labels.

We would like to use Prophecy\footnote{/examples/ACASX\_ProphecyTool.ipynb in the repository} to extract rules leading to each unique advisory (such as COC, Weak Left so on). The model does not predict a single class but five values; scores corresponding to each individual advisory. Therefore, a labels dataset is generated externally that contains for each input the respective class (horizontal advisory with minimum score). The command shown below invokes \textit{analyze} command with $type=3$ and the externally generated labels dataset. The rules are extracted after the layer \textit{relu\_3} in terms of the neuron values. Only the rule with the highest support for each label are stored in $ruleset.csv$.

\begin{small}

\noindent{\verb|$ python -m prophecy.main analyze|}

\verb|-m 'PROPHECY_PATH/dataset_models/acasx_onnx_keras.h5'|

\verb|-wd 'PROPHECY_PATH/results/acasx/rules/'|

\verb|-tx ./x_train_rshape.npy -ty ./y_predict_npy.npy|

\verb|-vx ./x_test_rshape.npy -vy ./y_predict_test_npy.npy|

\verb|-type 3 -layer_name 'relu_3' -top True|

\end{small}

The \textit{prove} command is invoked to verify the rule for label 0 (COC) with $pred=True$ and $min\_const=True$. This invokes Marabou with the following query 
for every $c' \neq 0$,
$\exists x \mid 
(F_{op}(x)[c'] < F_{op}(x)[0])$. The rule is Proved when all the above queries return UNSAT.\\

\begin{small}

\noindent{\verb|$ python -m prophecy.main prove|}

\verb|-m 'PROPHECY_PATH/dataset_models/acasxu/acasx_onnx_keras.h5'|

\verb|-wd 'PROPHECY_PATH/results/results/acasxu/rules/'|

\verb|-tx ./x_train.npy -label 0 -pred True -min_const True|

\verb|-mp '/usr/local/lib/python3.11/dist-packages/maraboupy/'|

\verb|-onx 'PROPHECY_PATH/dataset_models/|

\verb|   acasxu/ACASXU_experimental_v2a_1_1.onnx'|

\verb|-onx_map 'PROPHECY_PATH/dataset_models/|

\verb|   acasxu/acasx_onnx_map.csv'|

\end{small}

\subsection{Prophecy on a regression model:}
\label{subsec:reg}
We consider a regression model which takes in the following sensor inputs (i) Frequency, (ii) Airfoil angle of attack, (iii) Chord length, (iv) Free-stream velocity and (v) Suction side displacement thickness to estimate scaled sound pressure level. The model has a Mean Absolute Error (MAE) of 0.192 on the train set. 

Prophecy is used to infer rules corresponding to correct and incorrect model behavior, based on the post-condition property, $P(F(x)) := |F(x) - output_{ideal}| < MAE(0.192)$. As mentioned in Section~\ref{subsec:acts_col}, Prophecy currently handles post-condition properties not corresponding to prediction of classification models by setting the \textit{type} parameter to 3 and expecting the input labels dataset to directly contain the labels produced by the evaluation of $P$ on each input. Accordingly, the property is evaluated (prior to the invocation of Prophecy) on each train (and test) input to produce binary labels (1: satisfaction, 0: violation). This generates the labels datasets $y\_train.npy$ and $y\_test.npy$ respectively. The analyze command is then invoked with $type=3$ to extract rules corresponding to the binary labels in these datasets. The rules are conditions on neuron values of the specified layer.

\begin{small}

\noindent{\verb|$ python -m prophecy.main analyze|}

\verb|-m 'PROPHECY_PATH/dataset_models/airfoil_self_noise/model.h5'|

\verb|-wd 'PROPHECY_PATH/results/airfoil_self_noise/rules/'|

\verb|-tx ./x_train.npy -ty ./y_train.npy|

\verb|-vx ./x_test.npy -vy ./y_test.npy|

\verb|-layer_name 'layer_4_output_0' -type 3|

\end{small}

The rules corresponding to label 1 represent inputs on which the model displays correct behavior. In order to check if the property $P(F(X))$ is satisfied for all inputs satisfying $\sigma$, we would need to know the $output_{ideal}$ for every possible input in this region, which is not available. Therefore, we used a modified version of the post-condition property, $P'(F(X)):= ((min - 0.192) < F(X) < (max + 0.192))$, where $min$ and $max$ correspond to the minimum and maximum output values over all inputs in the train set satisfying $\sigma$. In other words, we try to check if the outputs for all inputs in the region corresponding to $\sigma$ fall within a desirable valid output region.

As mentioned in Section~\ref{subsec:acts_col}, for post-condition properties not corresponding to prediction of classification models, the $prove$ command can take in a constraints file whose path is specified in the $cp$ parameter. Each row in this file should adhere to the format $[output\_node\#,operator,value]$. Currently the implementation supports operators $MIN$ and $MAX$. For this example, $P'(F(X))$  is encoded using the following two rows, where 0 is the index of the output node of the network.
$$[0,MIN,0.192]$$
$$[0,MAX,0.192]$$ 
\\ 

\begin{small}

\noindent{\verb|$ python -m prophecy.main prove|}


\verb|-onx 'PROPHECY_PATH/dataset_models/|

\verb|   airfoil_self_noise/renamed_model.onnx'|

\verb|-onx_map 'PROPHECY_PATH/dataset_models/|

\verb|   airfoil_self_noise/airfoil_onnx_map.csv'|

\verb|-mp '/usr/local/lib/python3.11/dist-packages/maraboupy/'|

\verb|-wd 'PROPHECY_PATH/results/airfoil_self_noise/rules/'|

\verb|-tx ./x_train.npy -label 1|

\verb|-cp 'PROPHECY_PATH/dataset_models/airfoil_self_noise/rules/consts.csv'|

\end{small}


\subsection{Monitoring runtime inputs:}
\label{sec:monitor}

Prophecy is used to extract rules for correct vs incorrect classifications for models processing tabular data (Pima Diabetes (PD~\cite{PD})) and images (MNIST~\cite{MNISTWebPage}, CIFAR~\cite{CIFAR-10}). The command below shows the use of $-odl$ and $-oal$ parameters which extract rules after every dense and corresponding activation layers. 

\begin{small}

\noindent{\verb|$ python -m prophecy.main analyze|}

\verb|-m 'PROPHECY_PATH/dataset_models/mnist_cnn.h5'|

\verb|-wd 'PROPHECY_PATH/results/mnist/rules/'|

\verb|-tx ./x_train.npy -ty ./y_train.npy|

\verb|-vx ./x_test.npy -vy ./y_test.npy|

\verb|-odl -oal -type 1|

\end{small}

\begin{small}

The \textit{monitor} command is used at inference time to predict the model behavior on unseen inputs. For inputs that satisfy a larger number of rules for correct classification (vs incorrect) across layers, an output of \textit{correct} is produced, and an output of \textit{incorrect} is produced for those inputs satisfying more rules for incorrect classification. When there is a tie, an \textit{uncertain} output is produced. 

The original implementation of Prophecy executed the Extract module by default and stored the extracted rules in the form of a dictionary. However, searching through such a dictionary for all rules is expensive, especially for a runtime monitoring application. The \textit{rules} parameter (feature of  ProphecyPlus) uses this implementation.

\noindent{\verb|$ python -m prophecy.main monitor|}

\verb|-m 'PROPHECY_PATH/dataset_models/mnist_cnn.h5'|

\verb|-wd 'PROPHECY_PATH/results/mnist/rules/'|

\verb|-tx ./x_val.npy -ty ./y_val.npy rules|

\end{small}

\begin{small}

In the new implementation, ProphecyPlus, the decision-tree estimators are stored as is (the Extract step can also be skipped by setting the $sr$ flag). Invocation of the monitor command using the \textit{classifiers} parameter accesses the tree directly to predict the class for a runtime input.

\noindent{\verb|$ python -m prophecy.main monitor|}

\verb|-m 'PROPHECY_PATH/dataset_models/mnist_cnn.h5'|

\verb|-wd 'PROPHECY_PATH/results/mnist/rules/'|

\verb|-tx ./x_val.npy -ty ./y_val.npy classifiers|

\end{small}



\subsection{Prophecy on Clip - VLM Results}
\label{sec:vlm2}

In another experiment, we consider 4 cases. Case 1: images of \textit{car} misclassified, Case 2: images of \textit{car} correctly classified, Case 3: images of \textit{truck} correctly classified, Case 4: images of \textit{truck} misclassified. We assign labels \textit{0,1,2,3} to each of them respectively. Table \ref{tab:vlm2} shows detailed results of the experiments. 

\begin{table}[h!]
\begin{scriptsize}
\centering
\begin{tabular}{|*{14}{c|}}
\hline
\multirow{2}{*}{\textbf{Case}} & \multirow{2}{*}{\textbf{Neuron}}	& \multirow{2}{*}{\textbf{Sign.}}	& \multirow{2}{*}{\textbf{Supp.}}	& \multirow{2}{*}{\textbf{Cls.}}& 	\textbf{Train}& 	\textbf{Train}	& \textbf{Train}	& \textbf{Train}& 	\textbf{Test}	& \textbf{Test}	& \textbf{Test}& 	\textbf{Test}& 	\multirow{1}{*}{\textbf{Avg}} \\ 

&&&&&\textbf{Cov}&\textbf{Prec}&\textbf{Recall}&\textbf{F1}&\textbf{Cov}&\textbf{Prec}&\textbf{Recall}&\textbf{F1}&\textbf{F1} \\ \hline


	\multirow{6}{*}{1}&  396&$\leq$ -2.909   & 	\multirow{6}{*}{5}	&  \multirow{6}{*}{0}	&  \multirow{6}{*}{1.44}	& \multirow{6}{*}{100.00}	& \multirow{6}{*}{25.00}	& \multirow{6}{*}{40.00}	& \multirow{6}{*}{3.36}	& \multirow{6}{*}{20.00}& 	\multirow{6}{*}{4.76}& 	\multirow{6}{*}{7.69}& 	\multirow{6}{*}{23.85}  \\ 
& 188122&$\leq$ 6.002&&&&&&&&&&& \\
& 157846 &$\leq$ 4.471&&&&&&&&&&& \\
& 203801 &>2.509&&&&&&&&&&& \\
& 239132 &>1.152&&&&&&&&&&& \\ \hline 
    	    
\multirow{5}{*}{2} & 396 	& $\leq$ -2.909  & 	\multirow{5}{*}{114}	&  \multirow{5}{*}{1}	&  \multirow{5}{*}{32.76}	& \multirow{5}{*}{100.00}	& \multirow{5}{*}{76.00}	& \multirow{5}{*}{86.36}	& \multirow{5}{*}{29.53}	& \multirow{5}{*}{59.09}& 	\multirow{5}{*}{52.00}& 	\multirow{5}{*}{55.32}& 	\multirow{5}{*}{70.84}  \\ 
&188122 & $\leq$ 6.002&&&&&&&&&&& \\
&157846 & $\leq$ 4.471&&&&&&&&&&& \\
&203801& $\leq$ 2.509&&&&&&&&&&& \\
&88270& $\leq$ 4.828&&&&&&&&&&& \\ \hline 


\multirow{6}{*}{3} & 396,	&  >-2.909 & 	\multirow{6}{*}{117}	&  \multirow{6}{*}{2}	&  \multirow{6}{*}{33.62}	& \multirow{6}{*}{100.00}	& \multirow{6}{*}{78.00}	& \multirow{6}{*}{87.64}	& \multirow{6}{*}{30.87}	& \multirow{6}{*}{58.70}& 	\multirow{6}{*}{54.00}& 	\multirow{6}{*}{56.25}& 	\multirow{6}{*}{71.95} \\ 

& 33 &$\leq$-0.350 &&&&&&&&&&& \\
& 36996 &>-4.715&&&&&&&&&&& \\
& 154735&$\leq$3.052&&&&&&&&&&& \\
& 126911&$\leq$4.830&&&&&&&&&&& \\
& 92660&$\leq$2.879&&&&&&&&&&& \\ \hline 
 
\multirow{5}{*}{4}&  396, 	&  >-2.909 & 	\multirow{5}{*}{10}	&  \multirow{5}{*}{3}	&  \multirow{5}{*}{2.87}	& \multirow{5}{*}{100.00}	& \multirow{5}{*}{35.71}	& \multirow{5}{*}{52.63}	& \multirow{5}{*}{4.70}	& \multirow{5}{*}{42.86}& 	\multirow{5}{*}{10.71}& 	\multirow{5}{*}{17.14}& 	\multirow{5}{*}{34.89}  \\ 	    
& 33 & >-0.350&&&&&&&&&&& \\
& 219348 &>0.846&&&&&&&&&&& \\
& 4638&>-2.466&&&&&&&&&&& \\
& 153077&$\leq$1.175&&&&&&&&&&& \\ \hline 

\end{tabular}
\caption{Results for Running Prophecy on CLIP. }
\label{tab:vlm2}
\end{scriptsize}
\end{table}

\end{document}